\documentclass{article}
\usepackage{spconf,amsmath,amssymb,graphicx,hyperref}
\usepackage{comment}
\usepackage{subcaption} 

\title{Session-Level Spoken Language Assessment with a Multimodal Foundation Model via Multi-Target Learning}
\name{Hong-Yun Lin, Jhen-Ke Lin, Chung-Chun Wang, Hao-Chien Lu, Berlin Chen}
\address{Department of Computer Science and Information Engineering, National Taiwan Normal University}

\begin{document}
%
\maketitle
\begin{abstract}
Spoken Language Assessment (SLA) estimates a learner’s oral proficiency from spontaneous speech. The growing population of L2 English speakers has intensified the demand for reliable SLA, a critical component of Computer Assisted Language Learning (CALL). Existing efforts often rely on cascaded pipelines, which are prone to error propagation, or end-to-end models that often operate on a short audio window, which might miss discourse-level evidence. This paper introduces a novel multimodal foundation model approach that performs session-level evaluation in a single pass. Our approach couples multi-target learning with a frozen, Whisper ASR model-based speech prior for acoustic-aware calibration, allowing for jointly learning holistic and trait-level objectives of SLA without resorting to handcrafted features. By coherently processing the entire response session of an L2 speaker, the model excels at predicting holistic oral proficiency. Experiments conducted on the Speak \& Improve benchmark demonstrate that our proposed approach outperforms the previous state-of-the-art cascaded system and exhibits robust cross-part generalization, producing a compact deployable grader that is tailored for CALL applications.
\end{abstract}

\section{Introduction}
\label{sec:intro}

Spoken Language Assessment (SLA) aims to estimate a learner's oral proficiency from spontaneous speech, providing crucial support for Computer Assisted Language Learning (CALL) \cite{IEEE5881478}. Effective SLA requires the integration of evidence from multiple dimensions, including delivery, language use and content \cite{Peng_SLT2024_SAMAD,oh25_interspeech}. The complexity is further compounded by the diverse accents, disfluencies, and grammatical errors common among L2 speakers \cite{ma25b_interspeech,qian25_slate}.

The Speak \& Improve (S\&I) corpus and its associated 2025 Challenge provide a realistic benchmark for SLA, which includes multi-part, open-ended speaking tasks \cite{qian25_slate}. Each speaker’s session consists of Parts 1, 3, 4, and 5: P1 and P5 include multiple short responses. Human raters assign a single score to each part, even when it comprises several audio segments, and subsequently synthesize the evidence across all responses into a holistic proficiency score \cite{knill25_slate}. In contrast, most automated systems depart from this holistic procedure. Many competitive systems are limited to grading individual audio, training part-specific graders and subsequently fusing their scores in a post-hoc manner \cite{lin25_slate,cai25_slate}. While this strategy simplifies optimization, it results in fragmented modeling and may introduce error propagation across modules.

The evolution of SLA systems has progressed through several paradigms. Early work relied on statistical models with handcrafted features \cite{bernstein90_icslp,1997IEEE659144,ZechnerEvanini2019}. Subsequent systems leveraged text encoders on ASR transcripts \cite{qian25_slate,ZechnerEvanini2019} and later, self-supervised speech encoders like wav2vec \cite{lin25_slate,Baevski2020_wav2vec2}. While powerful, these cascaded pipelines are susceptible to error propagation and demand extensive, part-specific tuning \cite{qian25_slate,cai25_slate}. End-to-end models mitigate some of these issues but typically operate on short audio segments ($\leq$30 seconds), making it difficult to reason about discourse-level coherence \cite{ma25b_interspeech}.

This highlights a fundamental mismatch: human assessment is performed at the session level, whereas most models operate at the utterance level. Recent Multimodal Large Language Models (MLLMs) show promise, but they are often constrained to short audio inputs, complicating the evaluation of entire sessions that include multiple responses \cite{ma25b_interspeech,qian25_slate}. This work addresses these gaps by proposing a single-model MLLM that ingests a candidate's \emph{entire} session. It employs Multi-Target Learning (MTL) to predict all part-specific scores and an overall score in a single pass, aligning the automated process more closely with human evaluation. We also introduce an acoustic-aware calibration mechanism via a Whisper-based speech prior to enhance robustness. Our results demonstrate that this unified, session-level approach not only simplifies the assessment pipeline but also achieves state-of-the-art performance.

\begin{figure*}[!t]
  \centering
  \begin{minipage}[b]{0.47\linewidth}
    \centering
    \includegraphics[width=\linewidth]{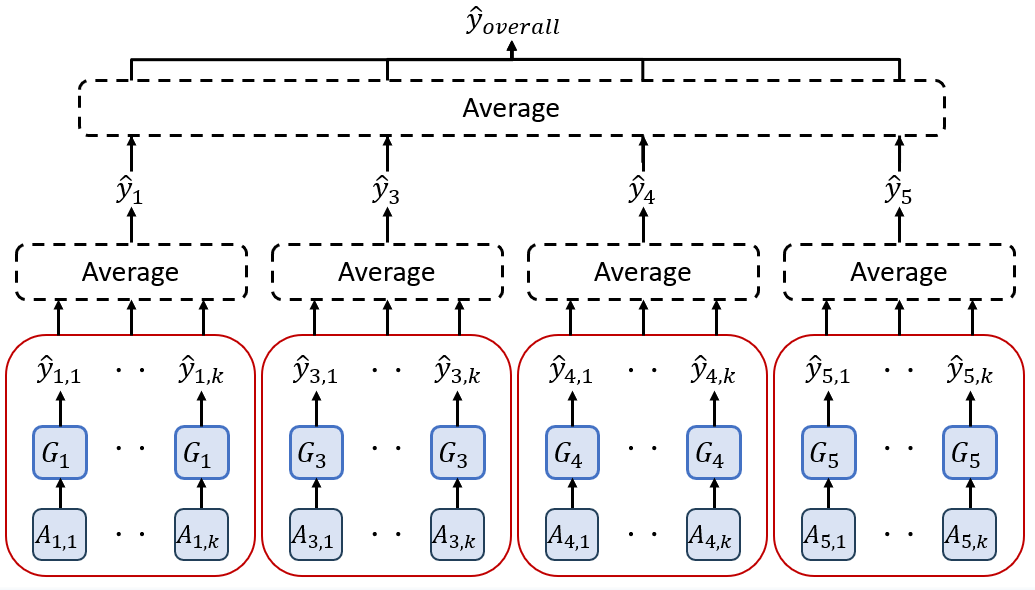}
    \vspace{2pt}
    \small (a) \textbf{Ensemble (Ens)}: per-part graders $G_p$ produce item-level predictions $\{\hat{y}_{p,i}\}$ that are averaged into a part score $\hat{y}_p$; the overall $\hat{y}_{\text{overall}}$ is the mean of the four part scores.
  \end{minipage}\hfill
  \begin{minipage}[b]{0.47\linewidth}
    \centering
    \includegraphics[width=\linewidth]{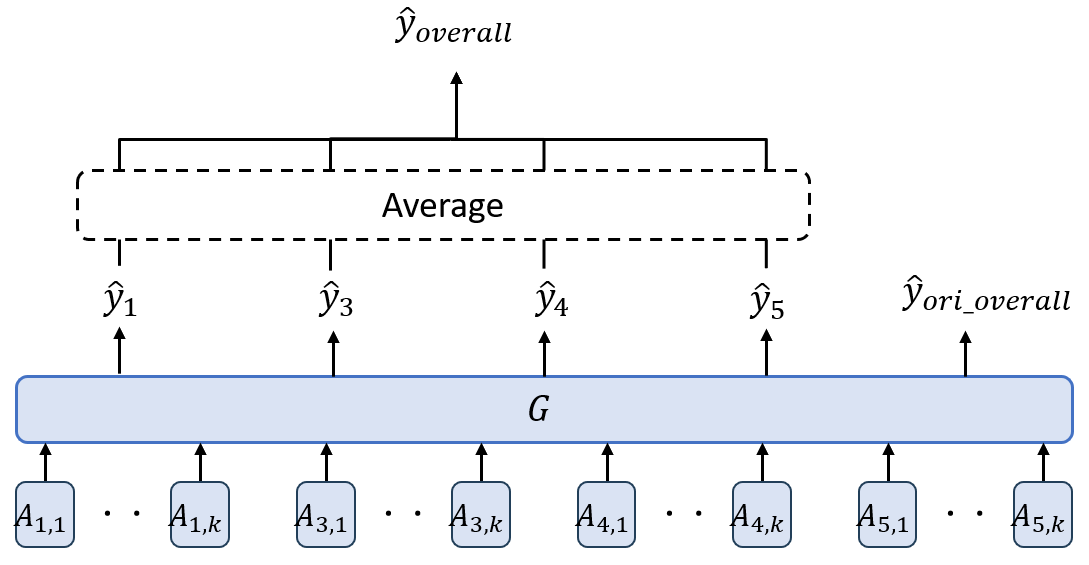}
    \vspace{2pt}
    \small (b) \textbf{Unified (Uni)}: a single session-level multimodal foundation model $G$ performs \emph{multi-target} regression, jointly emitting $(\hat{y}_{1},\hat{y}_{3},\hat{y}_{4},\hat{y}_{5})$ and an overall head $\hat{y}_{\text{ori\_overall}}$ in one pass; we also report the part-mean overall (average of the four part heads).
  \end{minipage}
  \caption{Ensemble (Ens) vs.\ Unified (Uni) session-level model designs}
  \label{fig:stg_vs_mtl}
\end{figure*}

\section{Methodology}
\label{sec:method}

\subsection{Problem Formulation and Modeling Goals}
\label{sec:method:problem}
We address \emph{session-level} spoken language assessment. A single test-taker provides multiple responses across parts $\mathcal{P}=\{\mathrm{P1},\mathrm{P3},\mathrm{P4},\mathrm{P5}\}$. The objective is a multi-target regression that maps the entire session to a five-dimensional score vector:
\[
\mathbf{s}=\bigl[s_{\mathrm{P1}},\,s_{\mathrm{P3}},\,s_{\mathrm{P4}},\,s_{\mathrm{P5}},\,s_{\mathrm{overall}}\bigr]\in\mathbb{R}^{5},
\]
where the first four components are part-level scores and the last is an overall score. As human raters consider all responses within a part before scoring \cite{knill25_slate,qian25_slate}, our approach is designed to perform reasoning within the model rather than relying on utterance-local predictions and external pooling.

Our design adheres to three primary goals:
\textbf{(i) Multimodal Integration:} Jointly exploit acoustic--prosodic and semantic--syntactic evidence, as proficiency judgments depend on delivery, language use, and content.
\textbf{(ii) Single-Model, Single-Pass Aggregation:} Aggregate evidence from the entire session within one model, aligning with human rating practices and mitigating error propagation from cascaded pipelines \cite{qian25_slate}.
\textbf{(iii) Parameter-Efficient Adaptation:} Enable practical fine-tuning for long, multi-utterance inputs and diverse L2 speech.

To achieve this, our model processes a unified multimodal sequence, allowing its attention mechanism to capture discourse-level dependencies. To further enhance its ability to model delivery, we introduce a speech-only branch that generates a Whisper-derived \emph{Acoustic Proficiency Prior} (APP), as detailed in \S\ref{sec:method:appfusion}.

\textbf{Prior Work} Our previous work on S\&I has explored two design lines: (i) \emph{Phi-4-STG}, an ensemble of per-part graders that captures part-specific cues but requires multiple large language models and post-hoc score fusion; and (ii) \emph{Phi-4-CTG} is a cross-task–generalized grader that processes each response separately: a single Phi-4 model is fine-tuned on pooled data from all parts, takes one audio segment at a time, and produces per-response scores that are averaged into part and overall results. As it lacks a multi-target head, CTG cannot capture discourse-level evidence across responses~\cite{lin25_slate}. These observations motivate a \emph{single-pass, session-level} alternative that performs \emph{multi-target} regression to jointly predict the four part scores and the overall score within one backbone. To supply non-lexical cues crucial for delivery (e.g., prosody and fluency) in a unified manner, we derive a \emph{Whisper-based Acoustic Proficiency Prior (APP)} from a frozen speech encoder and inject it as a learnable prior-prefix token into the multimodal sequence. This design preserves prosodic information while keeping modeling and deployment compact and unified within a multimodal foundation model.

Figure~\ref{fig:stg_vs_mtl} contrasts the prior ensemble (\emph{Phi-4-STG}) with our session-level design. The ensemble averages item-level predictions within each part and then across parts, but requires multiple large models and post-hoc fusion, leaving discourse-level aggregation outside the backbone. In contrast, a single multimodal model ingests all responses in one pass to jointly predict the four part scores and the overall score, internalizing discourse evidence and avoiding cross-module error propagation. For comparability with the ground truth, we additionally report the part-mean overall, since the dataset computes the overall label by averaging the four part scores.

\subsection{Model Architecture}
\label{sec:method:arch}
\begin{figure}[t]
  \centering
 \includegraphics[width=\linewidth]{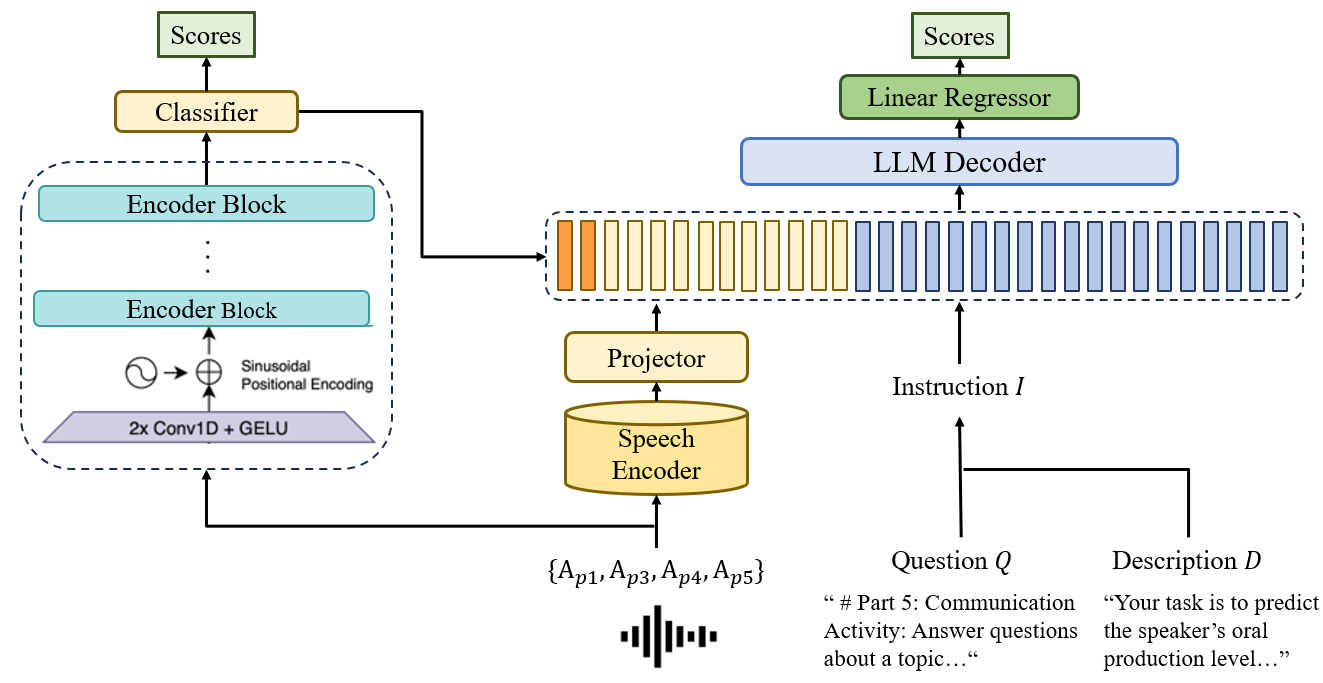}
  \caption{Overall architecture of the session-level MLLM grader.}
  \label{fig:arch}
\end{figure}

Building on the above observations from prior work, we adopt a single-pass session-level MLLM design that consolidates discourse-level aggregation and reduces cross-module error propagation. As depicted in Figure~\ref{fig:arch}, our architecture consists of a Phi-4 multimodal backbone for session-level aggregation and a parallel Whisper branch that yields an APP token, which is injected as a prefix into the backbone's input.

We use \emph{Phi-4-Multimodal} \cite{Phi4_TechReport_2024}, a generative MLLM. Its key advantage is a long context budget (128k tokens) and a dedicated speech pathway, making it suitable for robustly handling long-form, multi-utterance sessions without external chunking. A multi-output linear regression head is attached to the final hidden state to produce the five-dimensional score vector in a single pass.

Each session is formatted as a dialogue-style sequence, interleaving text-based instructions with audio responses. An audio placeholder token $\langle\!\lvert \mathrm{audio}_i \rvert\!\rangle$ is inserted for each response, and the corresponding 16\,kHz waveform is passed to the speech adapter. This design places lexical context and acoustic data into a shared attention space, retaining crucial prosodic information often lost in ASR-based pipelines.

\textbf{Whisper-derived Acoustic Proficiency Prior (APP)}
\label{sec:method:appfusion}
To explicitly capture non-lexical cues vital for proficiency assessment (e.g. fluency, hesitations), we introduce an \emph{Acoustic Proficiency Prior} (APP). This is motivated by findings that the acoustic realization of disfluencies is critical for robust speech understanding \cite{lin25b_slate} and complements the MLLM's focus on lexical content.

In parallel to the MLLM backbone, an acoustic prior is derived from a frozen Whisper model. Let $\mathbf{H}^{\mathrm{enc}}\!\in\!\mathbb{R}^{T_{\mathrm{enc}}\times d_w}$ denote the last hidden states of Whisper's encoder. After temporal mean pooling to obtain $\bar{\mathbf{h}}_{\mathrm{enc}}$, a two-layer MLP produces logits, followed by a softmax to yield the APP probabilities.
This probability vector captures speech-driven evidence complementary to text semantics. A projector maps the prior to a token embedding $\mathbf{e}_{\mathrm{prior}}$, which is prepended to the main multimodal sequence. This allows the model to perform session-level reasoning over a unified sequence that includes the APP token, serving as a reliability cue without hand-engineered late fusion.

\textbf{Prediction Head.} From the final hidden state $\mathbf{h}_T$, five scores are computed via a multi-output linear layer: $\hat{\mathbf{s}}=W\mathbf{h}_T+\mathbf{b}$. This latency-friendly parameterization preserves in-model aggregation.

\textbf{Parameter-Efficient Fine-Tuning (PEFT).} The model is adapted using LoRA \cite{hu2022lora}. With the backbone frozen, low-rank adapters are inserted into attention and MLP projections. Only these adapters and the regression head are optimized, enabling efficient MTL adaptation for long-context tasks.

\subsection{Learning Objective}
\label{sec:method:objective}
The primary loss function is a mean-squared error (MSE) over the five outputs. Let $y_{n,k}$ and $\hat{y}_{n,k}$ be the ground-truth and predicted scores for sample $n$ and component $k\in\{\mathrm{P1},\mathrm{P3},\mathrm{P4},\mathrm{P5},\mathrm{overall}\}$. The objective is to minimize:
\begin{equation}
\label{eq:mse}
\mathcal{L}=\frac{1}{\sum_{n,k} m_{n,k}}\sum_{n,k} m_{n,k}\,(\hat y_{n,k}-y_{n,k})^2
\end{equation}
This ensures that partially annotated samples contribute to training without introducing bias.

\section{Experiment}
\label{sec:experiment}

\subsection{Dataset}
We use the Speak \& Improve (S\&I) 2025 corpus, comprising roughly 315 hours of learner speech with part-level and overall holistic scores~\cite{knill25_slate}.
Each practice test has five parts in fixed order:
\emph{(P1) Interview}—8 short questions about the candidate (10\,s for the first four, 20\,s for the last four; the first two are unmarked and excluded);
\emph{(P2) Read Aloud}—8 sentences;
\emph{(P3) Long Turn 1}—a 1-minute opinion talk guided by three questions;
\emph{(P4) Long Turn 2}—a 1-minute description of a process shown in a graphic; and
\emph{(P5) Communication Activity}—five topic-related questions (up to 20\,s each).
The public release and our experiments focus on the open-speaking parts (P1/P3/P4/P5); P2 is not released~\cite{knill25_slate}.
For each session, human raters assign one holistic score to each open-speaking part; the dataset’s \emph{overall} label is defined as the arithmetic mean of these four part scores.

Targets are reported on the native S\&I scale aligned with CEFR-like bands
(2.0=A2, 2.5=A2+, 3.0=B1, 3.5=B1+, 4.0=B2, 4.5=B2+, 5.0=C1, 5.5=C1+).
We predict directly on this scale without post-hoc denormalization, and report all metrics on the original scale.

\subsection{Model Setup}
We fine-tune Phi-4-multimodal-instruct with the speech adapter enabled and attach a five-output linear head to jointly predict P1/P3/P4/P5 and overall in a single pass.
Inputs are dialog-style sequences that interleave part prompts with 16 kHz audio placeholders, allowing the backbone to attend across utterances within a session.
For the acoustic prior (APP), we use a frozen Whisper-large-v3 encoder \cite{Radford2022Whisper} to obtain last-layer representations, mean-pool them into a $D$-dimensional vector, and feed it to a two-layer MLP (Linear $D\!\to\!512$, GELU, Dropout 0.1, Linear $512\!\to\!8$).
The resulting prior is mapped to a token embedding and prepended to the multimodal sequence.
Training uses PEFT/LoRA for parameter efficiency.

Optimization uses AdamW with cosine decay and a brief warm-up:
learning rate $1{\times}10^{-4}$, weight decay $0.01$, warm-up $100$ steps, gradient clipping $1.0$, batch size $1$ with gradient accumulation $8$, for $3$ epochs.
FlashAttention~\cite{dao2024flashattention} is enabled.
The experimental settings and source code will be made publicly available in the camera-ready version.

\subsection{Evaluation Metrics}
At evaluation, we report per-target and overall Root Mean Squared Error (RMSE), Pearson correlation (PCC), and tolerance accuracies within $\pm 0.5$ and $\pm 1.0$ on the native scale.

\section{Results \& Discussion}
\label{sec:results}

\begin{table}[t]
\caption{Overall metrics on the S\&I evaluation set. 
``Arch'' denotes Architecture Type: 
Ens: ensemble (per-part models), Uni: unified (single model). 
For most models, RMSE is computed from a single holistic prediction head, 
whereas for MTL and MTL-APP it is derived from the average of four part-specific heads (P1, P3, P4, P5).}
\label{tab:tab2}
\resizebox{\columnwidth}{!}{%
\begin{tabular}{lccccc}
\hline
Arch & Model         & RMSE           & PCC            & \%$\le$0.5    & \%$\le$1.0    \\ \hline
Ens  & Perezoso      & 0.364          & 0.826          & 83.0          & \textbf{99.7} \\ \hline
Ens  & APP (Whisper)  & 0.383          & 0.805          & 81.7          & 99.0          \\
Ens  & Phi-4-STG     & 0.375          & 0.820          & 81.7          & 99.3          \\  \hline
Uni  & Phi-4-CTG     & 0.412          & 0.796          & 74.7          & 98.0          \\ 
Uni  & Phi-4-MTL     & 0.362          & 0.825          & \textbf{85.7} & 99.0          \\
Uni  & Phi-4-MTL-APP & \textbf{0.360} & \textbf{0.827} & \textbf{85.7} & 99.0          \\ \hline
\end{tabular}%
}
\end{table}

\begin{table}[t]
\caption{Per-part RMSE on the S\&I evaluation set (lower is better).}
\label{tab:tab1}
\centering
\resizebox{\columnwidth}{!}{%
\begin{tabular}{cccccc}
\hline
Model         & P1     & P3     & P4     & P5     & Overall \\ \hline
APP (Whisper)  & 0.581  & 0.461  & 0.497  &0.528 &0.383  \\
\hline
Phi-4-CTG     & 0.556  & 0.533  & 0.604  & 0.543  & 0.412   \\ 

Phi-4-MTL     & \textbf{0.494}  & 0.471  & 0.491  & 0.455  & 0.362   \\
Phi-4-MTL-APP & \textbf{0.494}  & \textbf{0.459}  & \textbf{0.485}  & \textbf{0.447}  & \textbf{0.360}   \\ \hline
\end{tabular}%
}
\end{table}

\begin{figure}[t]
  \centering
  \includegraphics[width=0.8\linewidth]{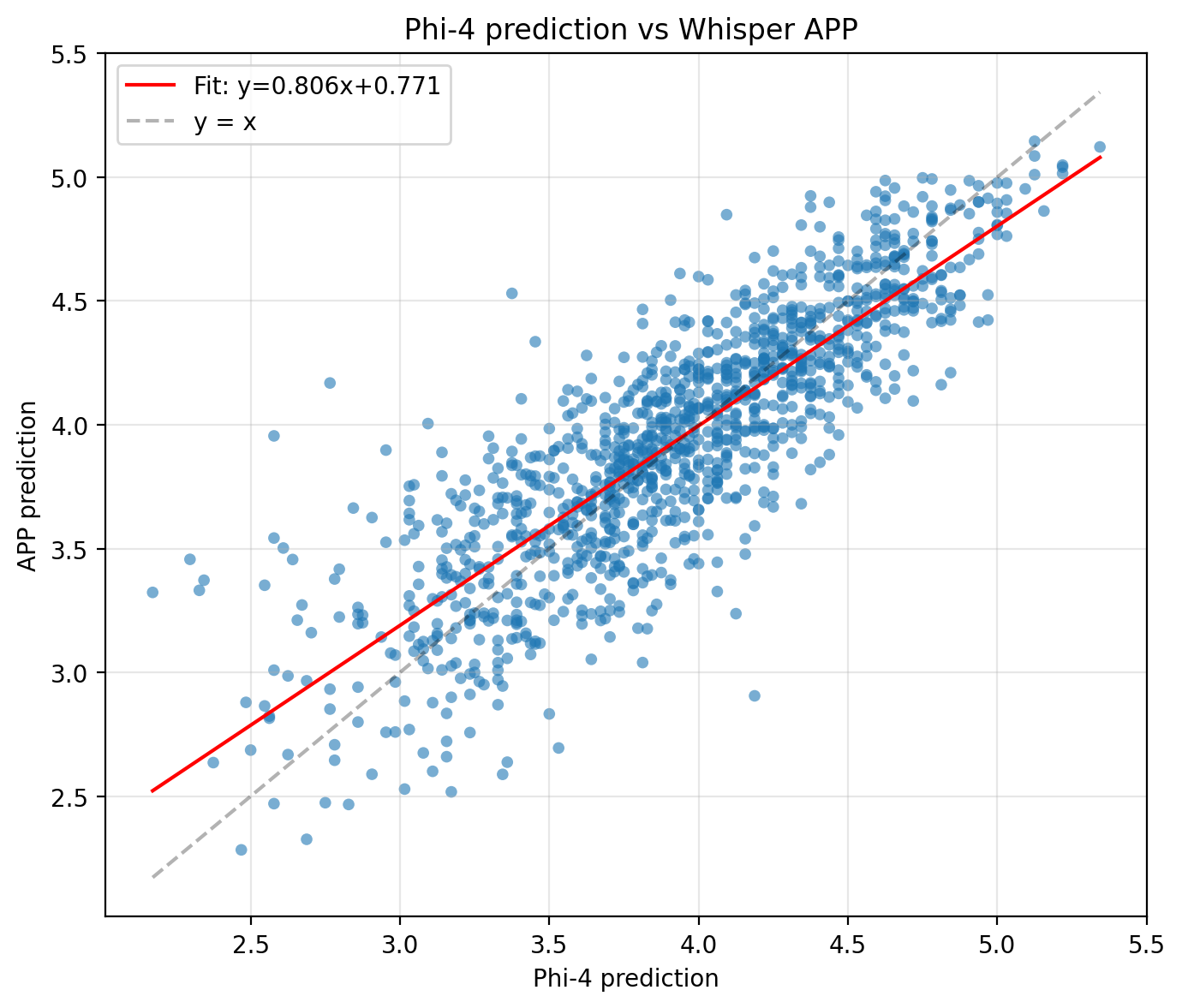}
  \caption{Regression scatter between APP(Whisper) and Phi-4 predictions for the overall score.}
  \label{fig:regression}
\end{figure}

The baseline systems in Tables~\ref{tab:tab2}--\ref{tab:tab1} span two architecture types: \emph{Ensemble (Ens)} and single-model \emph{Uni}. Include  \emph{Phi-4-STG} and \emph{Phi-4-CTG}, which were our submissions to the S\&I Challenge 2025 \cite{lin25_slate}.
We briefly highlight \emph{Perezoso}, the top leaderboard entry in the S\&I Challenge 2025 using a strong cascaded pipeline that fuses pretrained speech/text representations (e.g., Whisper and BERT) with a set of engineered prosodic and lexical features via linear regression, yielding competitive and interpretable performance while relying on external fusion and handcrafted features~\cite{cai25_slate}.

With a single session-level multimodal language model trained with multi-target learning (\emph{Phi-4-MTL}), we jointly predict all part-level scores and the overall score in one pass, leveraging cross-utterance and cross-part evidence while drastically reducing model complexity and training cost compared with ensemble models. As shown in Table~\ref{tab:tab2}, this single-model approach surpasses the per-part ensemble STG and the cascaded Perezoso, indicating that a unified session-level MLLM with multi-target learning can outperform state-of-the-art ensemble/cascaded systems while remaining compact and easier to deploy.

A closer comparison between MTL and CTG highlights the value of our approach. Although both are single-model setups,  MTL reduces error by more than 12\% relative and raises correlation. Table~\ref{tab:tab1} further shows that improvements are consistent across all parts, with clear gains on the multi-response parts (P1/P5). We attribute these gains to session-level aggregation: the model can reference multiple responses within the whole session, integrate delivery with content, and transfer information across different parts and responses, thereby addressing the discourse-level reasoning gap of CTG.

Finally, augmenting the unified grader with the Whisper-derived APP yields \emph{Phi-4-MTL-APP}, which provides consistent gains over MTL alone (Table~\ref{tab:tab2}). APP stabilizes delivery-sensitive cues such as fluency and pausing, improving calibration and accuracy. The benefit is particularly evident in the long-audio parts (P3/P4), where prosodic and pausing patterns are more informative, complementing the session-level reasoning already captured by the MLLM.

The regression plot in Fig.~\ref{fig:regression} further corroborates this effect: APP(Whisper) and Phi-4 predictions align closely, showing high correlation. This agreement indicates that the acoustic prior effectively supports the Phi-4 model as an acoustic-aware calibrator rather than introducing bias, and is especially helpful for the discourse-heavy long parts (P3/P4).

\section{Conclusion}
This paper presents a session-level SLA system based on a multimodal large language model with multi-target learning to jointly predict part-level and overall scores. A Whisper-derived Acoustic Proficiency Prior enhances calibration with delivery cues. On the Speak \& Improve benchmark, the unified approach achieves state-of-the-art performance (RMSE 0.360), consistently outperforming competitive ensemble and cascaded baselines, with clear gains on long and multi-response parts. Future work will extend the framework toward cross-task generalization and fairness considerations.

\vfill\pagebreak

\bibliographystyle{IEEEbib}
\bibliography{strings,refs}

\begin{thebibliography}{10}

\bibitem{IEEE5881478}
Wanyi Liu, Taogang Liu, Fengwen Liu, and Sumin Yang,
\newblock ``Computer assisted language learning,''
\newblock in {\em 2011 International Conference on E-Business and E-Government (ICEE)}, 2011, pp. 1--4.

\bibitem{Peng_SLT2024_SAMAD}
Wen-Hsuan Peng, Sally Chen, and Berlin Chen,
\newblock ``Enhancing automatic speech assessment leveraging heterogeneous features and soft labels for ordinal classification,''
\newblock in {\em 2024 IEEE Spoken Language Technology Workshop (SLT)}, 2024, pp. 945--952.

\bibitem{oh25_interspeech}
Sehyun Oh, Sunhee Kim, and Minhwa Chung,
\newblock ``{Multimodal and Multitask Learning for Predicting Multiple Scores in L2 English Speech},''
\newblock in {\em {Interspeech 2025}}, 2025, pp. 2425--2429.

\bibitem{ma25b_interspeech}
Rao Ma, Mengjie Qian, Siyuan Tang, Stefano Bannò, Kate~M. Knill, and Mark~J.F. Gales,
\newblock ``{Assessment of L2 Oral Proficiency using Speech Large Language Models},''
\newblock in {\em {Interspeech 2025}}, 2025, pp. 5078--5082.

\bibitem{qian25_slate}
{Mengjie Qian and Kate M. Knill and Stefano Bannò and Siyuan Tang and Penny Karanasou and Mark J.F. Gales and Diane Nicholls},
\newblock ``{Speak \& Improve Challenge 2025},''
\newblock in {\em {10th Workshop on Speech and Language Technology in Education (SLaTE)}}, {2025}, pp. {41--45}.

\bibitem{knill25_slate}
{Kate M. Knill and Diane Nicholls and Mark J.F. Gales and Mengjie Qian and Pawel Stroinski},
\newblock ``{Introducing the Speak \& Improve Corpus 2025: an L2 English Speech Corpus for Language Assessment and Feedback},''
\newblock in {\em {10th Workshop on Speech and Language Technology in Education (SLaTE)}}, {2025}, pp. {167--171}.

\bibitem{lin25_slate}
{Hong-Yun Lin and Tien Hong Lo and Yu Hsuan Fang and Jhen Ke Lin and Chung Chun Wang and Hao Chien Lu and Berlin Chen},
\newblock ``{The NTNU System at the S\&I Challenge 2025 SLA Open Track},''
\newblock in {\em {10th Workshop on Speech and Language Technology in Education (SLaTE)}}, {2025}, pp. {148--152}.

\bibitem{cai25_slate}
{Danwei Cai and Nitin Madnani and Kevin Yancey},
\newblock ``{Team Perezoso’s ASR and SLA System for Speak \& Improve Challenge 2025},''
\newblock in {\em {10th Workshop on Speech and Language Technology in Education (SLaTE)}}, {2025}, pp. {51--55}.

\bibitem{bernstein90_icslp}
Jared Bernstein, Michael Cohen, Hy~Murveit, Dimitry Rtischev, and Mitchel Weintraub,
\newblock ``Automatic evaluation and training in english pronunciation,''
\newblock in {\em First International Conference on Spoken Language Processing (ICSLP 1990)}, 1990, pp. 1185--1188.

\bibitem{1997IEEE659144}
Catia Cucchiarini, Helmer Strik, and Lou Boves,
\newblock ``Automatic evaluation of dutch pronunciation by using speech recognition technology,''
\newblock in {\em 1997 IEEE Workshop on Automatic Speech Recognition and Understanding (ASRU)}, 1997, pp. 622--629.

\bibitem{ZechnerEvanini2019}
Klaus Zechner and Keelan Evanini,
\newblock {\em Automated Speaking Assessment: Using Language Technologies to Score Spontaneous Speech},
\newblock Routledge, 2019.

\bibitem{Baevski2020_wav2vec2}
Alexei Baevski, Henry Zhou, Abdelrahman Mohamed, and Michael Auli,
\newblock ``wav2vec 2.0: A framework for self-supervised learning of speech representations,''
\newblock in {\em Advances in Neural Information Processing Systems 33 (NeurIPS 2020)}, 2020.

\bibitem{Phi4_TechReport_2024}
Marah Abdin, Jyoti Aneja, Harkirat Behl, S{\'e}bastien Bubeck, Ronen Eldan, Suriya Gunasekar, Michael Harrison, Russell~J Hewett, Mojan Javaheripi, Piero Kauffmann, et~al.,
\newblock ``Phi-4 technical report,''
\newblock {\em arXiv preprint arXiv:2412.08905}, 2024.

\bibitem{lin25b_slate}
{Jhen-Ke Lin and Hao-Chien Lu and Chung-Chun Wang and Hong-Yun Lin and Berlin Chen},
\newblock ``{Acoustically Precise Hesitation Tagging Is Essential for End-to-End Verbatim Transcription Systems},''
\newblock in {\em {10th Workshop on Speech and Language Technology in Education (SLaTE)}}, {2025}, pp. {163--166}.

\bibitem{hu2022lora}
Edward~J Hu, yelong shen, Phillip Wallis, Zeyuan Allen-Zhu, Yuanzhi Li, Shean Wang, Lu~Wang, and Weizhu Chen,
\newblock ``Lo{RA}: Low-rank adaptation of large language models,''
\newblock in {\em International Conference on Learning Representations}, 2022.

\bibitem{Radford2022Whisper}
Alec Radford, Jong~Wook Kim, Tao Xu, Greg Brockman, Christine McLeavey, and Ilya Sutskever,
\newblock ``Robust speech recognition via large-scale weak supervision,''
\newblock in {\em Proceedings of the 40th International Conference on Machine Learning}. 2023, vol. 202 of {\em Proceedings of Machine Learning Research}, pp. 28492--28518, PMLR.

\bibitem{dao2024flashattention}
Tri Dao,
\newblock ``Flashattention-2: Faster attention with better parallelism and work partitioning,''
\newblock in {\em The Twelfth International Conference on Learning Representations}, 2024.

\end{thebibliography}

\end{document}